\documentclass[10pt,twocolumn,letterpaper]{article}
\usepackage{cvpr} % To force page numbers, e.g. for an arXiv version

\usepackage{graphicx}
\usepackage{amsmath}
\usepackage{amssymb}
\usepackage{booktabs}
\usepackage{multirow}
\usepackage{bm}

\usepackage{hyperref}

\usepackage[capitalize]{cleveref}
\crefname{section}{Sec.}{Secs.}
\Crefname{section}{Section}{Sections}
\Crefname{table}{Table}{Tables}
\crefname{table}{Tab.}{Tabs.}

\begin{document}

\title{Causality is all you need}

\author{
    \@{Ning Xu$^{1}$, Yifei Gao$^{1}$, Hongshuo Tian$^{1}$, Yongdong Zhang$^{2}$, An-An Liu$^{1*}$} \and
    \@{$^1$ Tianjin University, China} \and
    \@{$^2$ University of Science and Technology of China, China} \and
    \@{\{ningxu,2022234109,kellyeden,liuanan\}@tju.edu.cn, \{zhyd73\}@ustc.edu.cn}
}
\maketitle

%%%%%%%%% ABSTRACT
\begin{abstract}
In the fundamental statistics course, students are taught to remember the well-known saying: ``Correlation is not Causation''.
Till now, statistics (i.e., correlation) have developed various successful frameworks, such as Transformer and Pre-training large-scale models, which have stacked multiple parallel self-attention blocks to imitate a wide range of tasks.
However, in the causation community, how to build an integrated causal framework still remains an untouched domain despite its excellent intervention capabilities.
In this paper, we propose the Causal Graph Routing (CGR) framework, an integrated causal scheme relying entirely on the intervention mechanisms to reveal the cause-effect forces hidden in data.
%eschewing the classic statistics method and instead
Specifically, CGR is composed of a stack of causal layers. Each layer includes a set of parallel deconfounding blocks from different causal graphs.
We combine these blocks via the concept of the proposed sufficient cause, which allows the model to dynamically select the suitable deconfounding methods in each layer.
CGR is implemented as the stacked networks, integrating no confounder, back-door adjustment, front-door adjustment, and probability of sufficient cause.
We evaluate this framework on two classical tasks of CV and NLP.
Experiments show CGR can surpass the current state-of-the-art methods on both Visual Question Answer and Long Document Classification tasks.
In particular, CGR has great potential in building the ``causal'' pre-training large-scale model that effectively generalizes to diverse tasks. It will improve the machines' comprehension of causal relationships within a broader semantic space.
%CGR has great potential in building the causal large-scale pre-training model, e.g., it can present the extensible platform to improve the machines' understanding of causal relations and shorten the process of realizing human-like way of thinking about the world.
\end{abstract}

%%%%%%%%% BODY TEXT
\section{Introduction}\label{sec:intro}

\emph{Correlation is not Causation.}

\quad\quad\quad\quad\quad\quad\emph{---Karl Pearson (1857 $-$ 1936)}
\\

In the fundamental statistics course, students are taught to remember the famous phrase: ``Correlation is not Causation". A classic example is the correlation between the rooster's crow and the sunrise. While the two are highly correlated, the rooster's crow does not cause the sunrise. However, statistics alone do not provide us what causation truly is. Unfortunately, many data scientists have a narrow focus on interpreting data without considering the limitations of their models. They mistakenly believe that all causal questions can be answered solely through data analysis and clever data-mining tricks.

Nowadays, thanks to the development of carefully crafted causal models \cite{pearl2018book, Peng000GYCZD00C22, CaiSL21, yang2021catt, yang2023deic}, the deep learning community has paid more and more attention to causation. Mathematically, the causal analysis aims to study the dynamic nature of distributions between variables. In statistics, we study and estimate various distributions and their model parameters from data, while in causal analysis, we study how when a change in the distribution of one variable affects the distribution of other variables. Definitedly, the change in the variable distribution is the \emph{do-operation}, which is an active \emph{intervention mechanism} in the data and can well define what is the causal effect between variables.

For example, given the environment $\mathcal{D}$, we take the input $X$ to predict the output $Y$, which is denoted as $P(Y|do(X), \mathcal{D})$, not $P(Y|X, \mathcal{D})$. The former represents the probability of $Y$ after $X$ is implemented on the pre-decision environment $\mathcal{D}$. The latter is the probability of $Y$ when $X$ coexists with the post-implementation environment $\mathcal{D}$. This coexistence environment may be different from the environment before the decision.
In short, statistics is observing something (i.e., seeing), and estimating what will happen. Causal analysis is an intervention, what is done (i.e., doing), and predicts what will happen \cite{pearl2018book}.

Till now, statistics have developed various successful frameworks, such as Transformer \cite{vaswani2017transformer}, Pre-training large-scale models \cite{dev2019bert, hao2019lxmert}, and so on. However, in the causation community, how to build an integrated causal framework still remains an untouched domain despite its excellent intervention capabilities. In this work, we propose the Causal Graph Routing (CGR), an integrated causal framework relying entirely on the intervention mechanisms to reveal the cause-effect forces hidden in data.
%eschewing the classic statistics method and instead

Specifically, the causal intervention aims to mitigate the effectiveness of confounding, which is a causal concept to describe the spurious correlation between input and output variables \cite{pearl2018book}. Because the noncausal paths are the source of confounding, we use the do-operator to control (or erase) the influence of noncausal paths, i.e., $P(Y|do(X))$, to deconfound $X$ and $Y$. Several classical deconfounding methods are presented in Fig.\ref{fig:sufficient+layers}:
%\textbf{a) Direct Effect}: The input $X$ has a direct causal effect on the output $Y$, i.e., $X\rightarrow Y$, where no confounder $Z$ exists.
\textbf{a) No Confounder}: The effect of $X$ on $Y$ via the mediator $M$, i.e., $X \rightarrow M \rightarrow Y$, where no confounder $Z$ exists.
\textbf{b) Back-door Adjustment}: The \emph{observable} confounder $Z$ influences both $X$ and $Y$, creating a spurious correlation $X\leftarrow Z\rightarrow Y$. The link $Z\rightarrow X$ is defined as the back-door path, which is blocked by controlling for $Z$.
\textbf{c) Front-door Adjustment}: The causal effect of $X$ on $Y$ is confounded by the \emph{unobservable} confounder $Z$ and linked by the mediator $M$. Furthermore, $M$ is observable and shielded from the effects of $Z$.
To eliminate the spurious correlation brought by $Z$, the front-door path, i.e., $M \rightarrow Y$, is blocked by controlling for $M$.

\begin{figure}[t]
  \centering
  
   \includegraphics[width=1\linewidth]{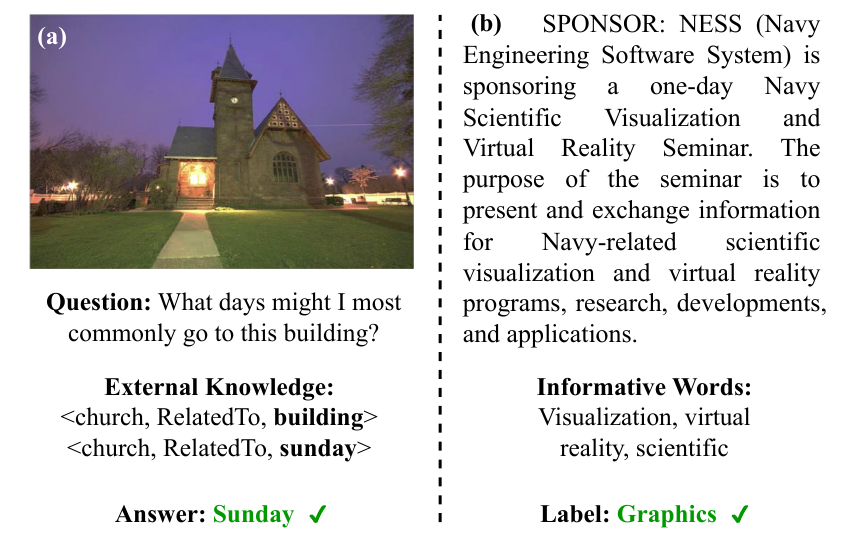}

   \caption{Two examples from (a) Visual Question Answering (VQA) task and (b) Long Document Classification (LDC) task.}
   \label{fig:draft}
\end{figure}

However, in several Computer Vision (CV) and Natural Language Processing (NLP) tasks, the causal intervention often requires the use of multiple deconfounding methods from different causal graphs.
%As shown in Fig.\ref{fig:draft}, in the Visual Question Answer (VQA) task\cite{ma2019okvqa}, to answer the question ``what days might I most commonly go to this building?'', the model first correctly detects ``building'' via the visual context, which may follow No Confounder or Front-door Adjustment (dataset bias). Then, the model correlates the object ``building'' with the fact $\langle church, RelatedTo, building\rangle$ and $\langle church, RelatedTo, sunday\rangle$ from the external knowledge base \cite{speer2017conceptnet}, which may be confounded with several irrelevant knowledge facts (language bias). This process need to be deconfounded by back- or front-door adjustments.
As shown in Fig.\ref{fig:draft}, in the Visual Question Answer (VQA) task\cite{ma2019okvqa}, to answer the question ``what days might I most commonly go to this building?'', the model first detects ``building'' via the visual context, which could be confounded by training data (dataset bias). To address this, the method like Front-door Adjustment or No Confounder is necessary.
Then, the model correlates the object ``building'' with the fact $\langle church, RelatedTo, building\rangle$ and $\langle church, RelatedTo, sunday\rangle$ from an external knowledge base \cite{speer2017conceptnet}, which could be confounded by irrelevant knowledge facts (language bias). To mitigate this, Back- or Front-door Adjustments are required for deconfounding.
Hence, relying on a single deconfounding method is insufficient to fulfill the requirement of deconfounding from diverse causal graphs. The same principle applies to the Long Document Classification (LDC) task\cite{dai2022trldc}.

Motivated by the transformer and its variants, which have stacked multiple parallel self-attention blocks to imitate a wide range of tasks \cite{yu2019mcan, dai2022trldc, hao2019lxmert}, we propose the Causal Graph Routing (CGR) framework, where above-mentioned deconfounding blocks are also stacked effectively.
Specifically, our framework is composed of a stack of causal layers. Each layer includes a set of parallel deconfounding blocks from different causal graphs.
We propose the concept of \emph{sufficient cause}, which provides the formal semantic for the probability that causal graph $A$ was a sufficient cause of another graph $B$.
It can chain together three candidates of deconfounding methods, i.e., no confounder, back-door adjustment, and front-door adjustment, to get the overall causal effect of $X$ on $Y$.
We calculate the weight of every deconfounding block to approximate the probability of sufficient cause, which allows the model to dynamically select the suitable deconfounding methods in each layer.
This facilitates the formulation of a causal routing path for each example.
CGR is implemented as the stacked networks that we assess on two classical tasks in CV and NLP.
Experiments show CGR outperforms existing state-of-the-art methods on both VQA and LDC tasks with less computation cost.
Notably, CGR exhibits significant potential for building the ``causal'' pre-training large-scale model, which can effectively generalize to diverse tasks. It will enhance the machines' understanding of causal relationships within a broader semantic space.
% We equip each module with a routing controller with XXX, which allows the model to dynamically select the suitable deconfounding methods in each layer,
% (so as to formulate the optimal routing path for each example.)

\section{Related Work}

\noindent \textbf{Causality.}
Causal inference \cite{pearl2018book} is an important component of human cognition.
Thanks to the development of carefully crafted causal methodologies, causality has been extensively studied and mathematized.
For examples,
Pearl et al. \cite{pearl2018book} propose the front- and back-door adjustments, which focus on removing unobservable or observable confounders by blocking noncausal paths.
Peng et al. \cite{Peng000GYCZD00C22} design the causality-driven hierarchical reinforcement learning framework.
%, instead of a randomness-driven exploration, to discover high-quality hierarchical structures.
Cai et al. \cite{CaiSL21} establish an algorithm to comprehensively characterize causal effects with multiple mediators.
Jaber et al. \cite{JaberRZB22} propose a new causal do-calculus for identification of interventional distributions in Partial Ancestral Graphs (PAGs).
These methods allow researchers to uncover the potential causal relationships between inputs and outputs to improve deep networks.
% and show it to be atomically complete.

\noindent \textbf{Applications on CV and NLP tasks.}
Cause-effect science is well suited for CV and NLP tasks.
For examples,
using the front-door adjustment to remove dataset bias for improving attention mechanisms \cite{yang2021catt}, discovering causal visual features on Video-QA task using the back-door adjustment \cite{zang2023discover},
equipping the pre-trained language model with a knowledge-guided intervention for text concept extraction \cite{YuanYLTLXX23},
%to mitigate spurious correlations
%using counterfactuals to measure the fairness in text classifiers \cite{GargPLTCB19},
generating counterfactual samples to mitigate language priors \cite{niu2021cfvqa}, and constructing a deconfounded framework for visual grounding \cite{huang2022devg} and image captioning \cite{yang2023deic}.

%The causal reasoning and debiasing methods have been widely applied in the field of visual language, such as
%using front-door adjustment to remove dataset bias to improve attention mechanisms \cite{yang2021catt}, discovering causal visual features related to questions in Video-QA tasks using back-door adjustment \cite{zang2023discover}, generating counterfactual samples to mitigate language priors in Visual-QA tasks \cite{niu2021cfvqa}, and constructing a deconfounded framework for visual grounding \cite{huang2022devg} and image captioning \cite{yang2023deic} tasks. Existing causal reasoning applications in the visual language domain follow the pipeline of ``identifying factors - constructing causal graphs - debiasing", while our work focuses on introducing multiple candidate causal graphs in the ``constructing causal graphs" step and allowing the model to adaptively select the appropriate causal graph for intervention.

% \begin{figure}[t]
%   \centering
  
%    \includegraphics[width=1\linewidth]{figs/causal layers.png}

%    \caption{Scheme of Causal Graph Routing.}
%    \label{fig:causal_layers}
% \end{figure}

\section{Methodology}

In this section, we discuss how to design the causal graph routing (Section \ref{subsec: cgr}), how to implement it into the stacked networks (Section \ref{subsec: stak cn}), and how to apply it with two classical CV and NLP tasks (Section \ref{subsec: appl}).

% The Causal Routing framework is illustrated in Figure 2. In the following sub-sections, we first introduce the causal routing construction and take VQA as a case study for multimodal tasks. We assume three possible causal relationships and construct corresponding framework of Transformer modules. Through causal routing, we discover the dominant causal models at both shallow and deep levels of the Transformer model.

\subsection{Causal Graph Routing}
\label{subsec: cgr}

To address the need of deconfounding from diverse causal graphs, we propose the causal graph routing framework, which can integrate different deconfounding blocks by calculating the probabilities of sufficient cause between causal graphs.
In particular, our objective is to dynamically select (routing) the suitable deconfounding methods for the given task.
We know three candidates of deconfounding methods which are \emph{no confounder}, \emph{back-door adjustment}, and \emph{front-door adjustment}.
For no confounder, we make the input $X$ to predict the output $Y$ via the mediator M without any confounder $Z$, denoted as $P_0\sim P(Y|do_0(X))$.
For back-door adjustment, we cut off the link $Z\rightarrow X$ to remove the spurious correlation caused by observable $Z$. It measures the average causal effect of $X$ on $Y$, denoted as $P_1\sim P(Y|do_1(X))$.
For front-door adjustment, we block the path $M\rightarrow Y$ by controlling for observable $M$, to remove the spurious correlation caused by unobservable $Z$, denoted as $P_2\sim P(Y|do_2(X))$.

\begin{figure}[t]
  \centering
  
   \includegraphics[width=\linewidth]{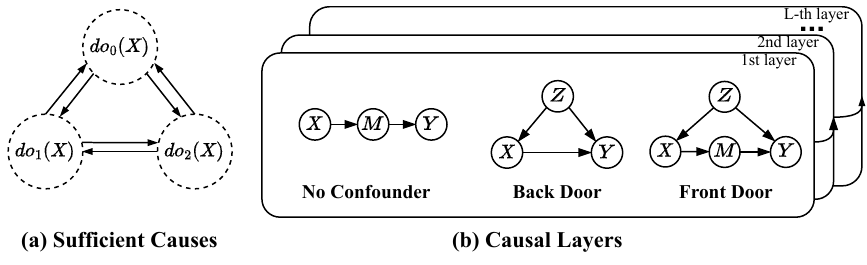}

   \caption{(a) Select the suitable causal graph (i.e., intervention operation) via the sufficient causes. (b) Scheme of Causal Graph Routing.}
   \label{fig:sufficient+layers}
   \vspace{-2.0ex}
\end{figure}

Intuitively, we can think of ``\emph{how to select the suitable causal graph}" as the game of building blocks. In this game, a modal is required to find the reasonable building method (i.e., causal graph) using the given units $X$, $Y$, $Z$, and $M$.
If the model finds the graph $P_1$ is unsuitable, it will switch its building method and consider using either the graph $P_2$ or $P_0$ instead.
Hence, there exists a hidden relevance among these graphs.
To formalize this relevance, we design the concept of \emph{sufficient cause} among graphs, which provides the formal semantic for the probability that causal graph $A$ was a sufficient cause of another graph $B$.
As shown in Fig.\ref{fig:sufficient+layers}(a), consider the arrow from $do_1(x)$ to $do_2(x)$ as an example, where the graph $P_1$ serves as a sufficient cause for the graph $P_2$.
We can represent the propositions $X=true$ and $Y=true$ as $x$ and $y$, respectively, while their complements are denoted as $x'$ and $y'$. The probability of sufficient cause from $P_1$ to $P_2$ is defined as:
\begin{equation}
\begin{split}
ps_{P_1\rightarrow P_2} = P( y_{do_2(x)} | y', do_1(x') )
\end{split}
\label{eq:causal routing arch}
\end{equation}
where $ps$ denotes the Probability of Sufficient cause that  measures the capacity of $do_1(x)$ to produce $do_2(x)$.
Given that the term ``production'' suggests a change from the absence to the presence of $do_2(x)$ and $y$, we calculate the probability $P(y_{do_2(x)})$ by considering situations where neither $do_1(x)$ nor $y$ are present.
In other words, $ps$ quantifies the effect of $P_1$ to cause $P_2$, which determines the probability of $y_{do_2(x)}$ occurring (the occurrence of $do_2(x)$ and $y$), given that both $do_1(x)$ and $y$ did not occur.
%In other words, $ps$ represents the probability of event $y_{do_2(x)}$ occurring (the occurrence of event $y$ in the absence of event $do_2(x))$, given that both $do_1(x)$ and $y$ did not occur.In a more coherent manner, $ps$ quantifies the ability of $P_1$ to cause $P_2$.It involves a transformation from the absence to the presence of $do_1(x)$. We condition the probability $P(y_{do_2(x)})$ on scenarios where both $y'$ and $do_1(x')$ are not present.$ps$ provides the probability that setting $P_1$ would result in the occurrence of $P_2$ when both $y'$ and $do_1(x')$ are absent.
Considering the sufficient causes from the other two graphs, the total effect (TE) of $P_2$ for the causal routing can be defined as:
\begin{equation}
\begin{split}
TE_{P_2} = P(y|do_2 (x)) *( ps_{P_1\rightarrow P_2} + ps_{P_3\rightarrow P_2} )
\end{split}
\label{eq:causal routing arch}
\end{equation}

We estimate the total effect of $X$ on $Y$, i.e., $P(Y|do(X))$, by dynamically routing all causal graphs:
\begin{equation}
\begin{split}
P(Y|do(X)) &= TE_{P_1} + TE_{P_2} + TE_{P_3} = \\
&P(y|do_1 (x)) *( ps_{P_2\rightarrow P_1} + ps_{P_3\rightarrow P_1} ) + \\
&P(y|do_2 (x)) *( ps_{P_1\rightarrow P_2} + ps_{P_3\rightarrow P_2} ) + \\
&P(y|do_3 (x)) *( ps_{P_1\rightarrow P_3} + ps_{P_2\rightarrow P_3} ) \\
\end{split}
\label{eq:causal_layer_eq}
\end{equation}
where $do(X)$ refers to the set of intervention operations, including $do_0(x)$, $do_1(x)$, and $do_2(x)$.
Till now, we have chained together the three deconfounding methods to get the overall causal effect of $X$ on $Y$. Each deconfounding method is equipped with two $ps$ terms, which stand for the probabilities of sufficient causes that another graphs would respond to the current graph.

As shown in Fig.\ref{fig:sufficient+layers}(b), we define the set of chained deconfounding methods as one \emph{casual layer}. We employ $L$ parallel casual layers, where each one produces the output values.
These values are then integrated to obtain the final values for the given task.

\subsection{Causal Stacked Networks}
\label{subsec: stak cn}

In this section, we illustrate how to implement our CGR in a deep framework. 
In practice, we adopt the stacked networks to perform the causal routing computation, integrating no confounder, back-door adjustment, front-door adjustment, and probability of sufficient cause. 

\subsubsection{Block of No Confounder} 

This block involves two stages: 1) extract the mediator $M$ from the input $X$ ($X \rightarrow M$) and 2) predict the outcome $Y$ based on $M$ ($M \rightarrow Y$). We have:

\begin{equation}
\begin{split}
    P(Y|X)=\sum_{m}P(M=m|X)P(Y|M=m)
\end{split}
\label{eq:direct prob}
\end{equation}

Considering that most CV and NLP tasks are formulated as classification problems, we compute $P(Y|X)$ using a transform function $f_{do_0}^{trans}(\cdot)$ and a multi-layer perceptron layer $\mathrm{MLP}(\cdot)$. The former aims to extract the mediator $M$ from the input $X$, while the latter outputs classification probabilities through the softmax layer.
\begin{equation}
\begin{split}
    \mathbb{E}_X(M)&=f_{do_0}^{trans}(X)\\
    P(Y|X)&= \mathrm{softmax} ( \mathrm{MLP}(\mathbb{E}_X(M)) )
\end{split}
\label{eq:direct network}
\end{equation}

\noindent We use the classical attention layer $\mathrm{Attention}(Q,K,V)\\=\mathrm{softmax}(Q K^T/\sqrt{d})V$  to calculate $f^{trans}_{do_0}(X) = \mathrm{Attention}(X,X,X)$. The queries $Q$, keys $K$ and values $V$ come from the input $X$. $d$ is the dimension of queries and keys. As shown in Fig.\ref{fig:causal_block}(a), in the $l$-th layer, we take the result of $\mathrm{MLP}(\mathbb{E}_X(M))$ as the output of the no confounder block $C^{(l)}_{do_0}$.

\subsubsection{Block of Back-door Adjustment}
We assume that an observable confounder $Z$ influences the relationship between input $X$ and output $Y$.
The link $Z\rightarrow X$ is blocked through the back-door adjustment, and then the causal effect of $X$ on $Y$ is identifiable and given by 
%(the detailed derivations are presented in supplementary material)
\begin{equation}
\begin{split}
    P(Y|do_1(X))=\sum_{z\in Z}P(Z=z)[P(Y|X, Z=z)]
\end{split}
\label{eq:back-door}
\end{equation} 
where $X$ and $Z$ denotes the embedding of inputs and confounders, respectively. To perform the back-door intervention operation, we parameterize $P(Y| X,Z)$ using a network, which final layer is a softmax function as:
\begin{equation}
\begin{split}
    P(Y | X, Z = z) = \mathrm{softmax}( f^{pre}_{do_1}( X, Z ) )
\end{split}
\label{eq:back-door network}
\end{equation}
where $f^{pre}_{do_1}(\cdot)$ is the fully connected layer predictor.
However, it requires an extensive amount of $X$ and $Z$ sampled from this network in order to compute $P(Y|do_1(X))$.
We employ the Normalized Weighted Geometric Mean (NWGM) \cite{xu2015show} to approximate the expectation of the softmax as the softmax of the expectation:
\begin{equation}
\begin{split}
P(Y|do_1(X))&=\mathbb{E}_{Z}( \mathrm{softmax} (f^{pre}_{do_1} (X,Z)) \\
&\approx \mathrm{softmax} (f^{pre}_{do_1} (\mathbb{E}_{X}(X), \mathbb{E}_{X}(Z)))
\end{split}
\label{eq:back-door network}
\end{equation}
We calculate two query sets from $X$ and $Z$ to estimate the input expectation $\mathbb{E}_{z}(X)$ and the confounder expectation $\mathbb{E}_{z}(Z)$, respectively, as:
\begin{equation}
\resizebox{0.88\linewidth}{!}{$
\begin{split}
    \mathbb{E}_{X}(X)&=\sum_{X=x}P(X=x|f_{ do_1 (X\rightarrow \mathbb{E}_{X}(X))}^{emb}(X))x\\
    \mathbb{E}_{X}(Z)&=\sum_{X=x}P(Z=z|f_{ do_1 (X, Z \rightarrow \mathbb{E}_{X}(Z))}^{emb}(X))x
\end{split}
\label{eq:back-door network}
$}
\end{equation}
where $f_{ do_1 (X\rightarrow \mathbb{E}_{X}(X))}^{emb}$ and $f_{ do_1 (X, Z \rightarrow \mathbb{E}_{X}(Z))}^{emb}$ denote query embedding functions.

\begin{figure}[tbp]
  \centering
  
   \includegraphics[width=\linewidth]{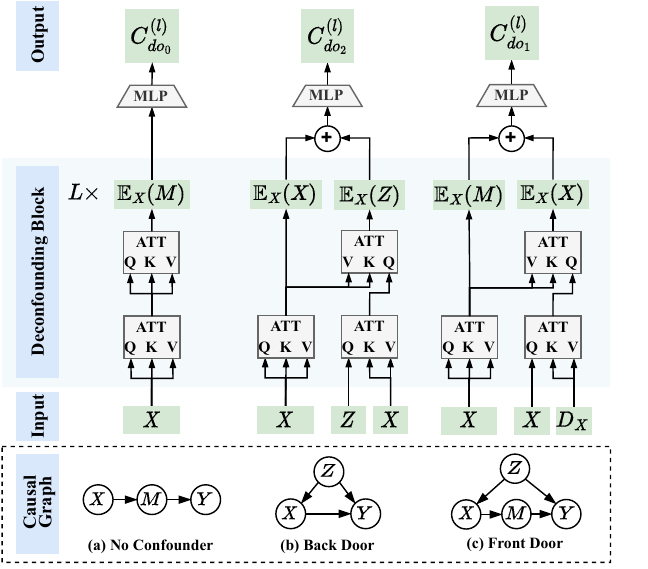}

   \caption{We adopt the stacked networks to perform the computation of causal graph routing.}
   \label{fig:causal_block}
   \vspace{-2.0ex}
\end{figure}

As shown in Fig.\ref{fig:causal_block}(b), we use the classical attention layer to estimate the expectations of both variables: $\mathbb{E}_X(X) = \mathrm{Attention}(X,X,X)$ and $\mathbb{E}_X(Z) = \mathrm{Attention}(Z,X,X)$. In the $l$-th layer, both of them are concatenated and passed through a multi-layer perceptron layer to produce the output of the back-door block $C^{(l)}_{do_1}$.

\subsubsection{Block of Front-door Adjustment}

We assume that an unobservable confounder $Z$ influences the relationship between input $X$ and output $Y$, while an observable mediator $M$ establishes a connection from $X$ to $Y$. We block the link $M\rightarrow Y$ through the front-door adjustment, and the causal effect of $X$ on $Y$ is given by 
%(the detailed derivations are given in supplementary material)

\begin{equation}
\resizebox{0.885\linewidth}{!}{$
\begin{split}
    P(Y|do_2(X))&=\\
    \sum_{M=m}P(&M=m|X)\sum_{x}P(X=x)P(Y|X=x,M=m)
\end{split}
$}
\end{equation}
\label{eq:front-door 2}

\noindent where $X$ and $M$ denotes the embedding of inputs and mediators, respectively.
Similar to the back-door adjustment, we parameterize $P(Y| X=x, M=m)$ using the softmax-aware network and the NWGM approximation. We have:
% We adopted the front-door adjustment approach, employing the NWGM approximation to incorporate the estimates of input X and mediator Z into the feature layer.
\begin{equation}
\resizebox{0.885\linewidth}{!}{$
\begin{split}
    % P(Y|do(X))&\approx softmax(f^{inf}_F(\hat{X}, \hat{Z}))\\
    P(Y |do_2(X)) &\approx softmax( f^{pre}_{do_2}( \mathbb{E}_{X}(X), \mathbb{E}_{X}(M) ) )
\end{split}
$}
\label{eq:front-door network}
\end{equation}

\noindent where $f^{pre}_{do_2}(\cdot)$ is the fully connected layer. 
Similarly, we estimate the expectations of variables by two query embedding functions $f_{ do_2 (X\rightarrow \mathbb{E}_X(X) )}^{emb}$ and $f_{ do_2 (X \rightarrow \mathbb{E}_X(M) )}^{emb}$.

\begin{equation}
\resizebox{0.88\linewidth}{!}{$
\begin{split}
    % P(Y|do(X))&\approx softmax(f^{inf}_F(\hat{X}, \hat{Z}))\\
    \mathbb{E}_X(X)&=\sum_{X=x}P(X=x|f_{ do_2 (X \rightarrow \mathbb{E}_X(X))}^{emb}(X))x\\
    \mathbb{E}_X(M)&=\sum_{M=m}P(M=m|f_{ do_2 (X \rightarrow \mathbb{E}_X(M))}^{emb}(X))m
\end{split}
\label{eq:front-door network}
$}
\end{equation}
% where $f_{F(X\rightarrow \hat{X})}^{emb}$ and $f_{F(X \rightarrow \hat{Z})}^{emb}$ represent the transformation network of input X into two query sets.

As shown in Fig.\ref{fig:causal_block}(c), we employ the classical attention layer to estimate the expectations, denoted as $\mathbb{E}_X(X) = \mathrm{Attention}(X,D_X,D_X)$ and $\mathbb{E}_X(M) = \mathrm{Attention}(X,X,X)$. 
Different from the above two blocks, we utilize the global dictionary $D_X$, which is initialized by conducting K-means clustering on all the sample features of the training dataset, to generate keys and values. 
In the $l$-th layer, the concatenated expectations are fed into a multi-layer perceptron to generate the output of the front-door block $C^{(l)}_{do_2}$.

\begin{figure}[t]
  \centering
  
   \includegraphics[width=1\linewidth]{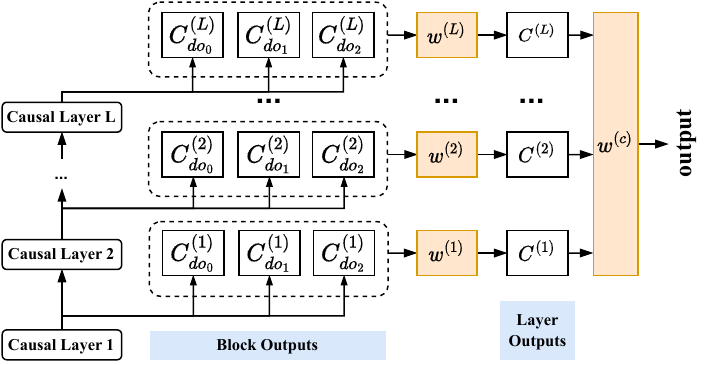}

   \caption{Routing architecture.}
   \label{fig:Routing}
   \vspace{-2.0ex}
\end{figure}

\subsubsection{Probability of Sufficient Cause}

In our framework, each layer consists of three deconfounding blocks from different causal graphs. 
We reexamine Eq.\ref{eq:causal_layer_eq}, and find that the process of evaluating the total effect $P(Y|do(X))$ is essentially to search the optimal deconfounding graph among three causal graphs, while the other two causal graphs are sufficient causal conditions for the optimal solution.
For example,
during the game of building blocks, if we discover that both graph $P_2$ and $P_0$ are ineffective, it naturally leads to use the graph $P_1$ as the optimal building method.
Take $TE_{P_1} = P_1(y|do_1(x)) *( ps_{P_2\rightarrow P_1} + ps_{P_3\rightarrow P_1} )$ an example.
When both $ps_{P_2\rightarrow P_1}$ and $ps_{P_3\rightarrow P_1}$ have high values, indicating that both $P_2$ and $P_3$ are sufficient cause for $P_1$, we consider $P_1$ as the optimal deconfounding method for achieving $P(Y|do(X))$.
Therefore, in this paper, we calculate the weight of the causal graph $P_1$ to approximate the probability of sufficient cause where $P_1$ is the optimal solution.
Similarly, we approximate the probabilities of sufficient causes for $P_2$ and $P_0$. 
%we calculate the weight of the causal graph $P_2$ to approximate the probability of $P_2$ being the optimal solution.
We have:
\begin{equation}
\begin{split}
C^{(l)} =   \sum^2_{i=0}  [f^{norm}( w^{(l)} )]_i  * C^{(l)}_{do_i}
\end{split}
\label{eq:causal weight 1}
\end{equation}
where $w^{(l)}$ is the weight vector of the $l$-th layer. Its each element $w^{(l)}_i (i=0,1,2)$ reflects the probability of the $i$-th deconfounding block as the optimal solution of the $l$-th layer. 
$f^{norm}(\cdot)$ denotes the normalization function (described in Optimization). 
$[\cdot]_i$ denotes the $i$-th element in a given vector. 
$C^{(l)}_{do_i}$ represents the output of the $i$-th deconfounding block in the $l$-th layer. 
$C^{(l)}$ is the output of the $l$-th causal layer. 
In this work, we employ $L$ parallel causal layers and combine them as: 
\begin{equation}
\begin{split}
C =  \sum^L_{l=1}  f^{norm}( w^{(c)} )  * C^{(l)}
\end{split}
\label{eq:causal weight 2}
\end{equation}
where $w^{(c)}$ is the layer-aware weight vector. $C$ represents the final output, which is computed as a weighted sum of all causal layers.
Both $w^{(l)}$ and $w^{(c)}$ are learnable parameters, initialized with equal constants, indicating that the routing weight learning starts without any prior bias towards a specific block or layer. 

\noindent \textbf{Stack.} 
In the no confounder block, the output expectation $\mathbb{E}_X(M)$ from the previous layer is used as the input $X$ for the current layer. In the back- and front-door adjustment blocks, the output expectation $\mathbb{E}_X(X)$ from the previous layer serves as the input $X$ for the current layer.

\noindent \textbf{Optimization.}
To enable the dynamic fusion of causal blocks and causal layers, we design the \emph{sharpening softmax function} to implement $f^{norm}(\cdot)$.
Specifically, we equip a temperature coefficient that converges with training for the ordinary softmax function as:
% Particularly, we remove the noise term $Gumbel(0,1)$ from the Gumbel softmax function as:
\begin{equation}
\begin{split}
    [f^{norm}(\alpha)]_i = \frac{exp(log(\alpha_i) / \tau)}{\sum_j exp(log(\alpha_j) / \tau)}
\end{split}
\label{eq:Optimization}
\end{equation}
where $\alpha$ represents the normalized weight vector after softmax; $\alpha_i$ is the $i$-th weight value; $\tau$ denotes the temperature coefficient for sharpening the softmax function. At the initial stage of training, the value of $\tau$ is set to 1, which results in the sharpening softmax function being the same as the regular softmax function. As the training progresses, $\tau$ gradually decreases, and as it converges to 0, the sharpening softmax function starts to resemble the argmax function more closely. By the designed sharpening softmax function, the block- and layer-aware weight vectors can be optimized through back-propagation. This optimization process enhances the performance of these weights, resulting in more noticeable differences after training.

\subsection{Application to Our Framework}
\label{subsec: appl}

\noindent \textbf{Visual Question Answering (VQA)} aims to predict an answer for the given question and image \cite{sta2015vqa}.
In this task, $X$ represents the input image-question pairs (e.g., an image involving ``church" and corresponding question ``what days might I most commonly go to this building?").
Y represents the output predicted answers (e.g., ``sunday"). 
$M$ is the mediator extracted from $X$, which refers to question-attended visual regions or attributes (e.g., a visual region involving ``church'' and an attribute ``building'').  %``yellow'' or ``green''
Additionally, in front-door adjustment, $Z$ denotes the unobservable confounder, while in back-door adjustment, $Z$ denotes the observable confounder that refers to question-attended external knowledge (e.g., $\langle church, RelatedTo, building\rangle$ and $\langle church, RelatedTo, sunday\rangle$). It is because external knowledge comprises both ``good'' language context and ``bad'' language bias \cite{niu2021cfvqa}. 
We use $L = 6$ parallel casual layers in VQA task. 
%It is because {\color{blue}{external knowledge represents the most probable response in the absence of visual information.}}
%Particularly, in back-door adjustment, we use the samples of image-question pair to represent $X$ []. 

\noindent \textbf{Long Document Classification (LDC)} aims to classify a given long document text \cite{dai2022trldc}.
In this task, $X$ represents the input document collection (e.g., legal-related documentation set).
$Y$ represents the output classification results (e.g., ``legal'' or ``politics'').
$M$ refers to segments extracted from document (e.g., a  segment ``...but in practice it too often becomes tyranny...'' indicates the label ``politics''). %some segments containing classification information, 
Similarly, $Z$ in front-door adjustment denotes the unobservable cunfounder, while $Z$ in back-door adjustment denotes the observable confounder that refers to the high-frequency words in each document. 
We use $L = 2$ parallel casual layers in LDC task. 
%{\color{blue}{It is because words with high TF-IDF values appear frequently in the text and are relatively rare in the entire corpus. These words are often more effective in distinguishing between different categories of document.}}
%Particularly, in back-door adjustment, $X$ represents the sample of document (i.e., a specific document).

The confounder extraction process for back-door adjustment is explained in Section \ref{sec:implement}.

\section{Experiment}

\subsection{Datasets and Metrics}

\noindent \textbf{VQA2.0} \cite{sta2015vqa}
is a widely-used benchmark VQA dataset, which uses images from MS-COCO.
It comprises a total of 443,757, 214,254, and 447,793 samples for training, validation, and testing, respectively.
Every image is paired with around 3 questions, and each question has 10 reference answers.
We consider both the soft VQA accuracy \cite{sta2015vqa} for each question type and the overall performance as the evaluation metrics.

\noindent \textbf{ECtHR} \cite{ilias2021ecthr}
is a popular dataset for the long document classification task.
It comprises European Court of Human Rights cases, with annotations provided for paragraph-level rationales.
The dataset consists of 11,000 ECtHR cases, where each case is associated with one or more provisions of the convention  allegedly violated.
The ECtHR dataset is divided into 8,866, 973, and 986 samples for training, validation, and testing, respectively.
It is used to evaluate the performance of our framework on a multi-label classification task.
Evaluation metrics include micro/macro average F1 scores and accuracy on the test set.

\noindent \textbf{20 NewsGroups} \cite{wahba2022svmplm}
is a popular dataset for the long document classification task.
It consists of approximately 20,000 newsgroup documents that are evenly distributed across 20 different news topics.
The dataset includes 10,314, 1,000, and 1,000 samples for training, validation, and testing, respectively.
It is used to evaluate the performance of our framework on a multi-class classification task.
We report the performance with the accuracy as the evaluation metric.

\label{sec:exp impl}

\subsection{Implementation}\label{sec:implement}

\noindent \textbf{Image and Text Processing.} For the VQA task, we employ the image encoder of BLIP-2 \cite{li2023blip2} to extract grid features.
We preprocess the question text and the obtained knowledge text in the back-door adjustment to lower case, tokenize the sentences and remove special symbols.
We truncate the maximum length of each sentence to 14 words, and utilize the text encoder of CLIP ViT-L/14 \cite{alec2021clip} to extract word features, followed by a single-layer LSTM encoder with a hidden dimension of 512.

For the LDC task, we define the maximum sequence length as 4,096 tokens. We split a long document into overlapping segments of 256 tokens. These segments have a 1/4 overlap between them. To extract text features, we utilize the pre-trained RoBERTa \cite{liu2019roberta} as the text encoder. %$\frac{1}{4}$

\noindent \textbf{Confounder Extraction.} 
For the VQA task, we use the question-attended external knowledge as the observable confounder in the back-door adjustment.
In this paper, we retrieve external knowledge from the ConceptNet knowledge base \cite{speer2017conceptnet}, which represents common sense using $\langle subject, relation, object \rangle$ triplets, such as $\langle church, RelatedTo, building \rangle$. 
Besides, ConceptNet provides a statistical weight for each triplet, ensuring reliable retrieval of information.
Specifically, we first extract 3 types of query words: (1) object labels of images obtained by GLIP \cite{li2022glip}; (2) OCR text of images by EasyOCR toolkit\footnote{https://github.com/JaidedAI/EasyOCR}; (3) n-gram question entity phrases.
All of words are filtered using a tool of part-of-speech restriction \cite{lin2019kagnet}, and the filtered words are combined to form the query set for searching common sense in ConceptNet.
We use the pre-trained MPNet \cite{song2020mpnet} to encode the returned common sense triplets and the given question.
Then, we calculate the cosine similarity between the encoded triplets and questions.
Further, the cosine similarity is multiplied by the given statistical weight to obtain the final score of a triplet.
We select the top-20 pieces of triplets as the observable confounder $Z$ for each image-question pair.

For the LDC task, we use the high-frequency words in each document as the observable confounder.
Specifically, we employ the TF-IDF method to select the top-M words in each long document ($M$ is set as 64 for ECtHR, and 128 for 20 NewsGroups). 
TF-IDF calculates the importance of a word within a document by considering its frequency in all documents.

\noindent \textbf{Training Strategy.} 
For the VQA task, we use the Adam optimizer to compute the gradient with an initial learning rate of $1\times 10^{-4}$, which decays at epoch 10, 12 with the decay rate of 0.5. 
We adopt a warm-up strategy for the initial 3 epochs, and the full model is trained for 13 epochs totally. 
The batch size is set to 64. 
For the LDC task, we use the AdamW optimizer with an initial learning rate of $2\times 10^{-5}$. 
We employ a linear decay strategy with a 10\% warm-up of the total number of steps to adjust the learning rate. 
We need about 16 epochs for the model to converge. 
The batch size on each GPU is set to 2. Our method is implemented on PyTorch with two 3090Ti GPUs.

\begin{table}[tp]
\caption{Comparison with the transformer-based models on VQA2.0.}
\centering
\resizebox{\linewidth}{!}{
\begin{tabular}{c|cccc|c}
\toprule[1.5pt]
\multirow{2}{*}{\textbf{Method}} & \multicolumn{4}{c|}{\textbf{Test-dev}}     & \textbf{Test-std} \\ \cline{2-6} 
                        & \textbf{overall} & \textbf{Yes/No} & \textbf{Num}   & \textbf{Others} & \textbf{overall}  \\ \midrule
Transformer\cite{vaswani2017transformer}             & 69.53   & 86.25  & 50.70 & 59.90  & 69.82    \\
DFAF\cite{gao2019dfaf}                    & 70.22   & 86.09  & 53.32 & 60.49  & 70.34    \\
ReGAT\cite{li2019regat}                   & 70.27   & 86.08  & 54.42 & 60.33  & 70.58    \\
MCAN\cite{yu2019mcan}                    & 70.63   & 86.82  & 53.26 & 60.72  & 70.90    \\
TRRNet\cite{yang2020trrnet}                  & 70.80   & -      & -     & -      & 71.20    \\
Transformer+CATT\cite{yang2021catt}        & 70.95   & 87.40  & 53.45 & 61.3   & 71.27    \\
AGAN\cite{zhou2020agan}                    & 71.16   & 86.87  & 54.29 & 61.56  & 71.50    \\
MMNAS\cite{yu2020mmnas}                   & 71.24   & 87.27  & 55.68 & 61.05  & 71.56    \\
TRAR$_S$\cite{zhou2021trar}                 & 72.00   & 87.43  & 54.69 & 62.72  & -        \\
TRAR$_S$(16*16)\cite{zhou2021trar}          & 72.62   & 88.11  & 55.33 & 63.31  & 72.93    \\ \midrule
CGR          & \textbf{75.46}   & \textbf{90.24}  & \textbf{57.16} & \textbf{67.01}  & \textbf{75.47}        \\ \bottomrule[1.5pt]
\end{tabular}
}
\label{tab:vqa sota}
\end{table}
\begin{table}[tp]
\caption{Comparison with the pre-trained large-scale models on VQA2.0.}
\centering
\resizebox{0.75\linewidth}{!}{
\begin{tabular}{c|c|c}
\toprule[1.5pt]
\textbf{Method}             & \textbf{Test-dev} & \textbf{Test-std} \\ \midrule
LXMERT\cite{hao2019lxmert}             & 72.42            & 72.54            \\
ERNIE-VIL\cite{yu2021ernie}          & 72.62            & 72.85            \\
UNITER\cite{chen2020uniter}             & 72.70            & 72.91            \\
12IN1\cite{lu202012}              & -                & 72.92            \\
LXMERT+CATT\cite{yang2021catt}        & 72.81            & 73.04            \\
LXMERT+CATT(large)\cite{yang2021catt} & 73.54            & 73.63            \\
VILLA\cite{gan2020villa}              & 73.59            & 73.67            \\
UNITER(large)\cite{chen2020uniter}      & 73.82            & 74.02            \\
VILLA(large)\cite{gan2020villa}       & 74.69            & 74.87            \\
ERNIE-VIL(large)\cite{yu2021ernie}   & 74.95            & 75.10            \\
 \midrule
CGR     & \textbf{75.46}            & \textbf{75.47}                \\ \bottomrule[1.5pt]
\end{tabular}
}
\label{tab:vqa pretrain}
\end{table}

\subsection{Results}

\noindent \textbf{Visual Question Answering.} We report the performance of our framework in VQA task against the transformer-based models (Tab.\ref{tab:vqa sota}) and the pre-trained large-scale models (Tab.\ref{tab:vqa pretrain}), in which test-dev and test-std are the online development-test and standard-test splits, respectively.
As shown in Tab.\ref{tab:vqa sota}, our CGR can significantly outperform all transformer-based models across all metrics.
Specially, our method outperforms the best competitor TRAR$_S$(16*16) by 3.91\% and 3.48\% under test-dev and test-std metrics, respectively,
which validates the effectiveness of the proposed stacked deconfounding method.
Meanwhile, the pre-training large-scale models also use the ``stacked'' mechanism, which stack multiple self-attention layers to imitate VQA task.
However, they ignore the negative impact of confounding that describes the spurious correlation between input and output variables.
Our method builds the multiple deconfounding layers to eliminate the spurious correlation. The better result in Tab.\ref{tab:vqa pretrain} showcase our advantage.

With less computation cost, our CGR has over 0.4\% improvement against the best competitor, ERNIE-VIL(large), in the pre-training large-scale models.
Remarkably, CGR equips with just 3 deconfounding methods with 6 causal layers, while ERNIE-VIL(large) relies on a much larger number of training attention layers (24 textual layers + 6 visual layers with 16 heads in each layer).
Besides, the causation community can offer numerous powerful deconfounding methods to further enhance our framework.
Hence, CGR has great potential for building the ``causal'' pre-training large-scale model to imitate a wide range of tasks.
This will greatly enhance machines' comprehension of causal relationships within a broader semantic space.

\begin{table}[]
\caption{Comparison with the state-of-the-arts on ECtHR and 20 NewsGroups.}
\centering
\resizebox{0.9\linewidth}{!}{
\begin{tabular}{c|c|cc}
\toprule[1.5pt]
                         & \multirow{2}{*}{\textbf{Method}}     & \multicolumn{2}{c}{$\bm{F_1}$}    \\ \cline{3-4} 
                         &                             & \textbf{Macro}         & \textbf{Micro}        \\ \midrule
\multirow{9}{*}{ECtHR}   & RoBERTa\cite{liu2019roberta}          & 68.9          & 77.3         \\
                         & CaseLaw-BERT\cite{zheng2021caselaw}      & 70.3          & 78.8         \\
                         & BigBird\cite{zaheer2020bigbird}          & 70.9          & 78.8         \\
                         & DeBERTa\cite{he2021deberta}        & 71.0          & 78.8         \\
                         & Longformer\cite{beltagy2020longformer}    & 71.7          & 79.4         \\
                         & BERT\cite{dev2019bert}   & 73.4          & 79.7         \\
                         & Legal-BERT\cite{ilias2020legalbert}  & 74.7          & 80.4         \\
                         & Hi-Transformer(RoBERTa)\cite{dai2022trldc} & 76.5          & 81.1         \\ \cline{2-4} 
                         & CGR & \textbf{76.6}          & \textbf{81.3}         \\ \toprule[1.5pt]
                         & \textbf{Method}                      & \multicolumn{2}{c}{\textbf{Accuracy}} \\ \midrule
\multirow{4}{*}{20 NewsGroups} & RoBERTa\cite{liu2019roberta}            & \multicolumn{2}{c}{83.8}     \\
                         & BERT\cite{dev2019bert}            & \multicolumn{2}{c}{85.3}     \\
                         & Hi-Transformer(RoBERTa)\cite{dai2022trldc}     & \multicolumn{2}{c}{85.6}     \\ \cline{2-4} 
                         & CGR & \multicolumn{2}{c}{\textbf{86.5}}     \\ \bottomrule[1.5pt]
\end{tabular}
}
\label{tab:ldc sota}
\end{table}

\noindent \textbf{Long Document Classification.} We compare the proposed framework with the state-of-the-art methods for LDC task on ECtHR and 20 News datasets (Tab.\ref{tab:ldc sota}).
Our method can consistently achieve better performance across all metrics, which indicates that our deconfounding strategy still works effectively on the challenging multi-class and multi-label NLP task.
Faced with the complex long texts, our CGR helps uncover potential cause effect and improve the model performance through multiple intervention routing. %operations.
Moreover, CGR can outperform Legal-BERT by 2.54\% under the Macro score. It suggests that our method has advantages in deconfouding the domain-specific knowledge.

\subsection{Ablation Studies}

We further validate the efficacy of the proposed framework by assessing several variants:
1) \textbf{One deconfounding block reserved per causal layer}: In our method, each causal layer consists of three deconfounding blocks. To verify their effectiveness, we retain only one block in each layer, i.e., no confounder, back-door adjustment, or front-door adjustment, respectively, and then calculate their average performance for comparison.
2) \textbf{Two deconfounding blocks reserved per causal layer}: Similarly, we retain two blocks in each layer, and calculate their average performance for comparison.
3) \textbf{Another strategy to calculate sufficient cause}:
We design the sharpening softmax function to calculate the weight of deconfounding block for the sufficient cause approximation. In this section, we remove the sharpening mechanism and adopt the ordinary softmax for Eq.\ref{eq:Optimization}, to obtain the sufficient cause.
Tab.\ref{tab: abla} reports the performance of ablation studies on VQA and LDC tasks.
Our framework outperforms all variants, which show the advantages of deconfounding from diverse causal graphs and our sufficient cause approximation method.

\begin{table}[]
\caption{Ablation studies on VQA2.0 and 20 NewsGroups.}
\centering
\resizebox{0.9\linewidth}{!}{
\begin{tabular}{c|c|c}
\toprule[1.5pt]
\textbf{Method}                     & \textbf{VQA2.0} & \textbf{20 NewsGroups} \\ \midrule
One deconfounding block reserved  & 69.35       & 84.40       \\
Two deconfounding blocks reserved & 70.03       & 85.00       \\
CGR w/o sharpen softmax    & 70.91       & 86.30       \\ \midrule
CGR                        & \textbf{71.20}       & \textbf{86.50}       \\ \bottomrule[1.5pt]
\end{tabular}
}
\label{tab: abla}
\end{table}

\subsection{Qualitative Analysis}

\begin{figure}[t]
  \centering
  
   \includegraphics[width=1\linewidth]{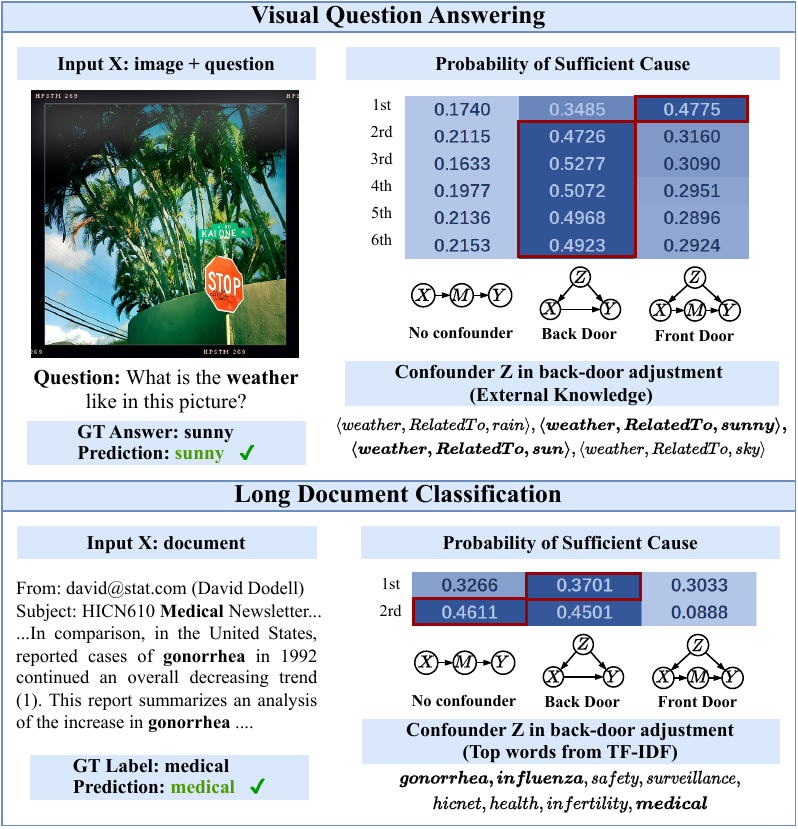}

   \caption{Qualitative examples from our method on VQA and LDC tasks. We present the probabilities of sufficient cause for all blocks in each layer and the corresponding confounder Z. The maximum value of each layer is highlighted with a red box.}
   \label{fig:Qualitative}
\end{figure}

Fig.\ref{fig:Qualitative} shows two qualitative examples from our method on VQA and LDC tasks.
To provide insight in which causal graph is dominant, we present the probabilities of sufficient cause in all layers, which reveal the explicit causal routing path within the framework.
See the first example, we observe that the front-door adjustment in the first layer dominates the answer inference, which helps the model avoid some unseen confounding effects, such as dataset bias.
As the routing progresses, the back-door adjustment has significantly enhanced, suggesting that the model start the focus on how to use external knowledge without confounding for the answer inference.

% {\color{blue}{Fig. XX shows two qualitative examples from our method on VQA and LDC tasks.
% To provide insight in which causal graph is dominant, we present the probabilities of sufficient cause in all layers, which reveal the explicit causal routing path within the framework.
% In case (a), for the VQA task, the CGR identifies the shallow reasoning process with the primary module being the front-door adjustment, while in the deep layers, the back-door adjustment becomes the main block. In the shallow layers of the Transformer, the front-door adjustment helps the model avoid some unseen confounding effects, such as dataset bias. In the deep layers of the Transformer, the back-door adjustment further enhances the model's understanding of crucial information regarding image-related questions. As shown in Figure 5(a), when answering the question "?", the extracted confounder includes "", enhancing the perception of information about the image region " ". }}

% For the LDC task, our CGR identifies the shallow reasoning process with the primary modules being the no confounder and back-door adjustment to mitigate confounding, while in the deep layers, the back-door adjustment becomes the dominant module. In the example shown in Figure 5(b), the extracted confounder enhances the hierarchical Transformer's attention to key classification information, such as "gonorrhea", "medical", "influenza", which are necessary causal conditions for the correct classification result "medical".

\subsection{Conclusion}

In this paper, we propose the novel Causal Graph Routing (CGR) framework, which is the first integrated causal scheme relying entirely on the intervention mechanisms to address the need of deconfounding from diverse causal graphs.
Specifically, CGR is composed of a stack of causal layers. Each layer includes a set of parallel deconfounding blocks from different causal graphs.
We propose the concept of sufficient cause, which chains together multiple deconfounding methods and allow the model to dynamically select the suitable deconfounding methods in each layer.
CGR is implemented as the stacked networks.
Experiments show our method can surpass the current state-of-the-art methods on both VQA and LDC tasks.
CGR has great potential for building the ``causal'' pre-training large-scale model. We plan to extend CGR with more powerful deconfounding methods and apply it into other tasks for revealing the cause-effect forces hidden in data.

%%%%%%%%% REFERENCES
{\small
\bibliographystyle{ieee_fullname}
\bibliography{egbib}
}

\end{document}